\documentclass[11pt]{article}
\usepackage[preprint]{acl}
\usepackage{times}
\usepackage{latexsym}
\usepackage[T1]{fontenc}
\usepackage[utf8]{inputenc}
\usepackage{CJKutf8}
\usepackage{microtype}
\usepackage{inconsolata}
\usepackage{graphicx}
\usepackage{booktabs}
\usepackage{amsmath}
\usepackage{amssymb}
\usepackage{dsfont}
\usepackage{algorithm}
\usepackage{algorithmic}
\usepackage{multirow}
\usepackage{colortbl}
\usepackage{etoc}
\usepackage{alaya-title}
\definecolor{openrow}{HTML}{EAF1FB}
\definecolor{openhdr}{HTML}{D8E6F7}
\definecolor{proprow}{HTML}{FDF1EA}
\definecolor{prophdr}{HTML}{FADFCE}
\definecolor{refrow}{HTML}{E9F6F0}
\definecolor{refhdr}{HTML}{D5EEE2}
\definecolor{headrow}{HTML}{EFF1F4}

\newcommand{\zh}[1]{\begin{CJK*}{UTF8}{gbsn}#1\end{CJK*}}
\newcommand{\huan}{{\CJKfamily{bsmi}嬛\CJKfamily{gbsn}}}
\newcommand{\bench}{ZhenhuanBench}
\newcommand{\method}{CPR-GRPO}
\newcommand{\agvr}{AGVR}
\title{Honeyed Words, Hidden Daggers: Benchmarking and Training LLMs to See Through Mock Politeness}
\AlayaAuthor[1,2,*]{Caixin Kang}
\AlayaAuthor[2,*]{Sitong Gong}
\AlayaAuthor[2,*]{Tianyu Yan}
\AlayaAuthor[2]{Kaipeng Zhang}
\AlayaAuthor[1]{Yoichi Sato}
\AlayaAuthor[1,2,\ensuremath{\dagger}]{Yifei Huang}
\AlayaAffiliation[1]{The University of Tokyo}
\AlayaAffiliation[2]{Alaya Lab}
\AlayaContribution[*]{Equal contribution}
\AlayaContribution[\ensuremath{\dagger}]{Corresponding author}
\AlayaInfo{Project page}{\url{https://ut-vision-social.github.io/}}
\AlayaInfo{Email}{\AlayaEmail{cxkang@iis.u-tokyo.ac.jp}, \AlayaEmail{hyf015@gmail.com}}
\AlayaInfo{Date}{October 12, 2026}
\hypersetup{pdfauthor={Caixin Kang, Sitong Gong, Tianyu Yan, Kaipeng Zhang, Yoichi Sato, Yifei Huang}}

\newcommand{\JudgeTThreeRho}{0.985}
\newcommand{\JudgeOverallRho}{0.999}

\hypersetup{pdftitle={Honeyed Words, Hidden Daggers: Benchmarking and Training LLMs to See Through Mock Politeness},colorlinks=true,citecolor=darkblue,linkcolor=darkblue,urlcolor=darkblue}
\AlayaAbstract{The Chinese idiom \zh{口蜜腹剑} (``honeyed words, hidden daggers'') names speech that is polite in form but hostile in function. We present \textbf{ZhenhuanBench}, a 1{,}592-item benchmark built from expert commentary on \emph{Empresses in the Palace}, covering speaker intent, emotion--language mismatch, and per-listener uptake. Text-only evaluation covers 34 LLMs. GPT-6 Astra leads the leaderboard at 73.48\% overall, while its listener-reaction score is 58.37\%. The annotations also supply supervised traces and string-verifiable training rewards. With 338 gold traces, an 8B model outperforms teacher distillation on identical items. Subsequent RL saturates, and chance correction reduces the estimated recoverable performance gap by 7.5$\times$. Our \textbf{CPR-GRPO} loop (Compress--Probe--Replenish) uses this estimate to weight gold supervision, reaching 67.0 on public multiple-choice tasks and 74.5 under the training-matched hidden protocol. Without target-task training, CPR outperforms Base, SFT, and GRPO across six Chinese and English pragmatic and sarcasm evaluations. On three external datasets, training with CPR yields higher mean Macro-F1 than SFT and GRPO.}
\begin{document}
\maketitle
\etocdepthtag.toc{mtmain}

\section{Introduction}
\label{sec:intro}

\begin{figure*}[t]
\centering
\includegraphics[width=\textwidth]{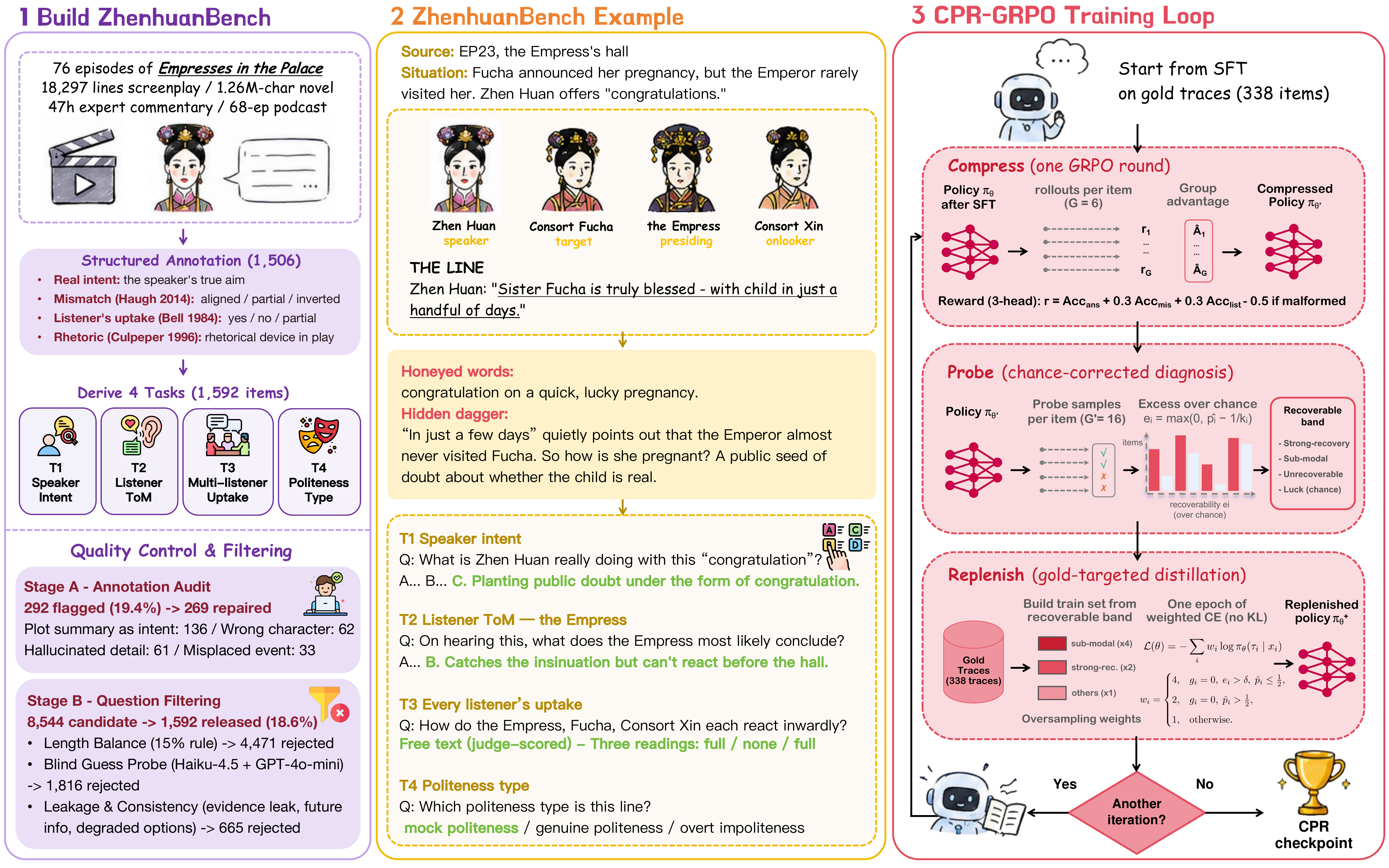}
\caption{
Overview of ZhenhuanBench and CPR-GRPO. (1) 47 hours of expert commentary become 1{,}506 structured annotations and four tasks, with 81.4 percent of candidate items rejected by quality gates. (2) A sample scene with multi-listener context translated into derived tasks T1–T4. (3) Training that reuses annotations as rewards and traces through Compress, Probe, and Replenish iterations (\S\ref{sec:cpr}).
}
\label{fig:overview}
\vspace{-0.2cm}
\end{figure*}

In \emph{Empresses in the Palace} (\zh{甄\huan{}传}), a consort congratulates a newly pregnant rival with a single warm line, \zh{「到底是富察姐姐有福气，三五日间就有喜了。」} (``Sister Fucha is truly blessed, with child in just a handful of days.''). The words are impeccably kind, yet everyone in the hall knows the Emperor has scarcely visited her, so the congratulation quietly asks how the child could be his. The Empress catches the barb and cannot smile; a less astute consort takes it as sincere praise. This is mock politeness, a pragmatic register in which politeness is the weapon rather than the message \citep{culpeper1996towards,haugh2014im}. Reading it requires recovering the speaker's true intent, noticing that emotion and linguistic surface point in opposite directions, and tracking which listeners decode the attack and which take it at face value \citep{bell1984language}. Figure~\ref{fig:overview} shows how a single expert-anchored annotation captures all three layers, and how the same structure later becomes training signal.

We study whether LLMs understand mock politeness, whether this ability can be trained into a small model, and how the training signal contributes to its performance. Existing resources provide limited scale and annotation detail for these questions. The largest prior Chinese mock-politeness evaluation contains 100 items with three-way labels and no theory-of-mind dimension \citep{maskofcivility2026}, and the classic Chinese implicature benchmark is nearly saturated by strong models \citep{yue2024swordsman}.

Our answer to the first question is \bench{}, a 1{,}592-item benchmark whose gold labels derive from 47 hours of expert commentary aligned to the original screenplay. Each aligned utterance carries the speaker's real intent, an emotion--language mismatch label, and per-listener uptake judgments, and from one annotation we derive four tasks probing intent, second-order theory of mind, multi-listener uptake, and Culpeper-style politeness classification. Text-only evaluation spans 34 LLMs. GPT-6 Astra leads the leaderboard with 73.48 overall and 58.37 on listener reactions; Gemini 3.7 Flash leads the MCQ tasks with 78.97 (\S\ref{sec:eval}). Providing expert commentary raises Gemma-4-31B's hidden-set accuracy from 65.7 to 85.2.

Our training study asks how to use these expert interpretations to train mock-politeness understanding. Annotation-Grounded Verifiable Rewards (\agvr{}) turn the annotated relations into supervision for an 8B model.
With 338 traces and LoRA, an 8B model rises from 36.8 to 55.7 on the leaderboard's MCQ tasks under supervision plus one GRPO round, above GPT-5, and gold traces beat chain-of-thought distilled from Kimi-K2.6 on identical items. Diagnosing why further RL saturates, we find the gap between pass@$k$ and greedy decoding shrinks 7.5$\times$ after correction for multiple-choice chance; the set of recoverable items decreases with each GRPO round. This diagnosis yields \method{}, a Compress--Probe--Replenish loop that directs further supervision from existing gold traces toward recoverable errors. The 8B model reaches 67.0 on the leaderboard\textquotesingle{}s MCQ tasks, and 74.5 under the training-matched hidden protocol. The trained model also improves over Base on Chinese and English sarcasm tests and retains its aggregate accuracy under character renaming. In summary, our contributions are as follows:

\begin{itemize}
\item We construct \bench{} (\S\ref{sec:benchmark}), the largest theory-grounded Chinese mock-politeness benchmark, whose expert-anchored annotations each yield four tasks over one scene, carry graded multi-source provenance, and double as training signal.
\item We develop \method{} (\S\ref{sec:agvr}, \S\ref{sec:cpr}) to train mock-politeness understanding from limited expert annotations, combining judge-free \agvr{} rewards with chance-corrected selection and weighting of existing gold traces. With 338 traces, the 8B model reaches 67.0 on the leaderboard's MCQ tasks.
\item We present extensive experiments and findings (\S\ref{sec:eval}, \S\ref{sec:training}). Text-only evaluation covers 34 LLMs, including controlled reasoning comparisons. Training controls show that gold annotation outperforms teacher distillation and interface-only SFT enables subsequent GRPO gains; chance correction reduces the estimated recoverable gap by 7.5$\times$.
\end{itemize}

\section{Related Work}
\label{sec:related}

\paragraph{Chinese pragmatics benchmarks.} SwordsmanImp evaluates sitcom implicature and is nearly saturated \citep{yue2024swordsman}, the Mask of Civility targets mock politeness directly but is small, coarse-grained, and unreleased \citep{maskofcivility2026}, and broader suites such as PUB lack the multi-listener dimension \citep{sravanthi2024pub}. LLM pragmatic competence is uneven and tuning-sensitive \citep{hu2023fine,ruis2023goldilocks}, and computational politeness predicts perceived politeness from surface features \citep{danescu2013computational} rather than asking whether a polite surface hides hostile intent. \bench{} is an order of magnitude larger than the closest mock-politeness resource, adds per-listener uptake and mismatch structure, and is designed to be trainable (Table~\ref{tab:prior}, Appendix~\ref{app:schema}).

\paragraph{Theory of mind and social intelligence.} Machine ToM is usually probed with synthetic false-belief vignettes \citep{le2019revisiting,kim2023fantom,gandhi2023understanding}, on which strong models remain brittle \citep{sap2022neural}, while broader benchmarks test social commonsense or interactive episodes \citep{sap2019socialiqa,zhou2024sotopia} and multimodal work reads deception, first impressions, and multi-speaker structure from what is visible and audible \citep{kang2025can,kang2026perception,ouyang2025multi}. Related systems span interactive digital humans \citep{cai2025interactive}, social video generation \citep{ouyang2026socialdirector}, narrative agents \citep{huang2025living}, and proactive assistance with temporal memory \citep{gong2026grove,gong2026vinci2}. \bench{} instead isolates the case where the signal is the language itself, with second-order questions from natural dialogue rather than constructed vignettes, several co-present listeners, and expert-documented rather than stipulated gold.

\paragraph{Distillation and RLVR.} Rationale-based training uses human explanations \citep{camburu2018esnli} and model-generated reasoning \citep{hsieh2023distilling,zelikman2022star}; our traces are instead programmatic rewrites of human-anchored annotation (\S\ref{sec:interface}). RLHF trains models from human preferences \citep{ouyang2022training}, while RLVR methods such as GRPO use verifiable outcomes, chiefly in mathematics and code \citep{shao2024deepseekmath,deepseekr1}, with recent work on social coordination \citep{kang2026pairedworlds}; \agvr{} uses exact matches against annotation fields to score pragmatic judgments. \citet{yue2025does} show that RLVR sharpens rather than expands ability, which we replicate and refine with chance correction. Our GRPO adopts DAPO's zero-variance filtering and asymmetric clip \citep{yu2025dapo}. Where variance curricula reweight by raw variance \citep{vcrl2025}, off-policy guidance mixes gold into rollout groups \citep{luffy2025}, and self-distillation retrains on model outputs \citep{gulcehre2023rest}, our loop assigns additional gold supervision to examples in a chance-corrected recoverable band. Chance correction is a classical psychometric tool \citep{cohen1960coefficient}; to our knowledge, no prior work has used it to gate RL training.

\section{\bench{}}
\label{sec:benchmark}

\subsection{Expert Commentary to Structured Gold}
\label{sec:pipeline}

We construct gold labels by extracting and auditing interpretations from expert commentary, with corroborating evidence retrieved from an independent English recap podcast and the source novel. Our primary source is the first 110 parts (47 hours of audio) of an ongoing video commentary series on \emph{Empresses in the Palace} by a dedicated analyst \citep{shulan2025xipin}, whose close readings of individual lines provide exactly the intent-level analysis the task needs (source selection and reception in Appendix~\ref{app:schema}). We transcribe the commentary with Whisper \citep{radford2023robust}, align commentary passages to the 18{,}297-utterance screenplay (1{,}214 scenes), and extract 1{,}506 alignments, each pairing an utterance with the expert's reading.

Cross-source corroboration uses two additional textual sources. \emph{Chasing Dramas}, an independent English recap podcast \citep{chasingdramas2019podcast}, contributes 68 episode transcripts (about 249k words), and the source novel by Liu Lianzi \citep{liulianzi2007hougong} contributes 1.26M characters including inner monologue. We index both sources with BGE-M3 \citep{chen2024bge} and retrieve passages for each alignment to compare the commentary's interpretation with the podcast's account and the novel's narration.

An LLM extractor converts each alignment into four structured fields with verbatim quote anchoring, and a separate audit pass checks fidelity to the commentary (Appendix~\ref{app:schema}). \texttt{real\_intent} states what the speaker is actually doing; \texttt{emotion\_language\_mismatch} labels affect against surface as \emph{aligned}, \emph{partial\_mismatch}, or \emph{inverted}, operationalizing implicational mismatch \citep{haugh2014im}; \texttt{third\_party\_reactions} lists each listener present with an uptake judgment (\emph{yes}, \emph{no}, \emph{partial}) and a one-sentence reason, operationalizing audience design \citep{bell1984language}; and \texttt{rhetoric\_sub\_type} supports Culpeper-style classification \citep{culpeper1996towards}.

\subsection{Tasks}
\label{sec:tasks}

From each alignment we derive up to four items. \textbf{T1} asks the speaker's true intent (4-way MCQ), the implicature-recovery problem of \citet{grice1975logic}. \textbf{T2} asks what a named listener privately concludes (4-way MCQ), a second-order theory-of-mind judgment \citep{premack1978does,perner1985john}. \textbf{T3} asks for free-text descriptions of every listener's reaction (scored by an LLM judge at evaluation only), operationalizing audience design \citep{bell1984language}. \textbf{T4} asks for a three-way politeness classification, \emph{genuine politeness}, \emph{mock politeness}, or \emph{overt impoliteness}, the face-based contrast of politeness theory \citep{brown1987politeness} operationalized through \citet{culpeper1996towards}. Deriving multiple tasks from one annotation enables cross-task consistency and per-alignment diagnostics unavailable to independent-item benchmarks (\S\ref{sec:diag}). Item construction enforces an anti-leakage checklist, with neutral phrasing, scene truncation before consequences unfold, length-balanced options, and three option permutations per MCQ (Appendix~\ref{app:schema}).

\begin{table*}[t]
\centering
\small

\setlength{\tabcolsep}{3.5pt}
\begin{tabular}{lccccc@{\hskip 10pt}lccccc}
\toprule
\rowcolor{headrow} Model & All & T1 & T2 & T3 & T4 & Model & All & T1 & T2 & T3 & T4 \\
\midrule
\rowcolor{openhdr} \multicolumn{12}{@{}l}{\emph{Open-weight models}} \\
Kimi K3 & 66.8 & 72.4 & 67.9 & 53.4 & 77.4 & Qwen3-8B (base) & 34.2 & 41.4 & 34.2 & 26.9 & 32.1 \\
DeepSeek V4 Pro & 62.5 & \textbf{76.2} & 58.8 & 49.4 & 66.0 & GLM-4.6V-Flash & 33.2 & 33.9 & 33.8 & 28.8 & 38.7 \\
Kimi-K2.6 & 62.0 & 66.1 & 65.4 & 48.2 & 71.7 & Qwen3.5-4B & 32.9 & 38.5 & 35.8 & 15.8 & 47.2 \\
Gemma-4-31B & 59.6 & 64.4 & 62.9 & 42.0 & 75.5 & InternVL3-8B & 31.8 & 29.7 & 32.5 & 26.0 & 46.2 \\
DeepSeek-V4-Flash & 56.8 & 67.8 & 52.1 & 45.3 & 65.1 & GLM-4.1V-9B & 31.1 & 30.5 & 35.8 & 25.8 & 32.1 \\
Qwen3-VL-235B & 52.4 & 56.1 & 49.2 & 42.6 & 70.8 & Qwen3-VL-4B & 30.3 & 32.2 & 32.5 & 22.0 & 36.8 \\
GLM-4.6V & 47.4 & 53.1 & 46.2 & 37.4 & 56.6 & GLM-4.5-Air & 30.1 & 28.5 & 28.7 & 29.5 & 37.7 \\
Qwen3-VL-30B & 44.4 & 53.1 & 41.2 & 36.9 & 46.2 & MiniCPM-o-2.6 & 29.4 & 28.0 & 31.2 & 21.7 & 43.4 \\
Qwen3-32B & 44.1 & 51.0 & 43.8 & 34.4 & 48.1 & MiMo-VL-7B & 28.3 & 25.9 & 29.2 & 28.7 & 31.1 \\
Qwen3.5-9B & 41.6 & 47.3 & 36.2 & 31.1 & 61.3 & Qwen2.5-VL-7B & 28.3 & 31.4 & 30.4 & 22.6 & 27.4 \\
Qwen3-14B & 39.1 & 46.9 & 34.2 & 30.9 & 49.1 & Qwen3.5-2B & 26.0 & 32.2 & 27.5 & 12.6 & 34.9 \\
Qwen3-VL-8B & 37.4 & 36.4 & 35.8 & 32.0 & 53.8 &  &  &  &  &  &  \\
\midrule
\rowcolor{prophdr} \multicolumn{12}{@{}l}{\emph{Proprietary models}} \\
GPT-6 Astra & \textbf{73.5} & 73.2 & \textbf{79.6} & \textbf{58.4} & \textbf{89.6} & GPT-5 & 52.3 & 52.3 & 51.7 & 49.5 & 59.4 \\
Gemini 3.7 Flash & 72.3 & 74.5 & 78.8 & 53.3 & \textbf{89.6} & GPT-4o & 49.3 & 51.0 & 50.4 & 39.0 & 63.2 \\
Qwen3.8-Max & 62.5 & 62.8 & 64.2 & 49.8 & 83.0 & Haiku-4.5 & 44.4 & 44.8 & 45.0 & 40.3 & 50.0 \\
Sonnet-4.6 & 58.9 & 59.4 & 60.4 & 50.6 & 70.8 & GPT-5-mini & 38.1 & 35.1 & 37.5 & 35.7 & 50.9 \\
Gemini-3.1-Flash-Lite & 58.6 & 66.1 & 57.1 & 45.8 & 69.8 & GPT-4o-mini & 30.8 & 26.8 & 29.6 & 28.3 & 47.2 \\
\midrule
\rowcolor{refrow} \method{} (Qwen3-8B) & 56.7 & 66.1 & 67.5 & 27.4 & 67.9 & \multicolumn{6}{l}{\footnotesize Trained reference; MCQ accuracy: 67.0} \\
\bottomrule
\end{tabular}
\caption{Public-set leaderboard (percent): 33 models and a trained 8B reference. Qwen3.8-Max judges the models' free-text T3 responses. All is the mean score over all 791 items. Within each group, models rank by All down the left block, then the right. Best scores across all models are bold.}
\label{tab:leaderboard}
\vspace{-0.2cm}
\end{table*}

\subsection{Splits and Scale}
\label{sec:splits}

\bench{} contains 1{,}592 items, split into 791 public, 319 hidden-test, and 482 training items with episode-level isolation between training and hidden episodes (Figure~\ref{fig:composition}; per-task-by-split counts in Appendix~\ref{app:stats}). Sampling ran in two symmetric passes over the full commentary to avoid coverage bias. T4 is the scarcest task (212 items) because it requires a confident Culpeper class from the commentary, while T3 contributes 429 free-text items that no MCQ-only protocol would cover. Training and hidden sets share no episode. All splits use the same drama and cast.

\begin{figure}[t]
\centering
\includegraphics[width=\columnwidth]{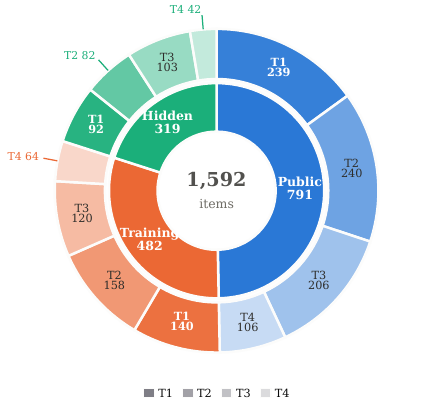}
\caption{\bench{} composition (1{,}592 items). Inner ring, the three splits (training and hidden episode-disjoint); outer ring, tasks T1--T4 shaded dark to light. Full counts in Appendix~\ref{app:stats}.}
\label{fig:composition}
\vspace{-0.2cm}
\end{figure}

\subsection{Quality}
\label{sec:quality}

During construction, every MCQ passes an options-only guessing filter using Claude Haiku-4.5 or GPT-4o-mini: the probe attempts the item from the answer options alone, with the scene withheld, and any item it can solve without the scene is rewritten, removing answer-selection artifacts in the spirit of adversarial dataset filtering \citep{lebras2020adversarial,gururangan2018annotation}. Across two assembly passes the item writer proposed 8{,}544 candidates and the automated checks rejected 81.4 percent, leaving the 1{,}592 final items (full funnel in Table~\ref{tab:funnel}).

Each item records its primary source, supporting quotations, and any cross-source corroboration. The gold label is anchored in the expert commentary for all 1{,}592 items, verbatim in 823 of them and as a supported paraphrase in the remaining 769, and 1{,}047 items (65.8 percent) additionally carry a quoted secondary passage that corroborates the same reading, 1{,}023 citing the English podcast and 28 the novel, with 4 items citing both. A \texttt{gt\_strength} field grades this support from 1 to 3 (79, 1{,}303, and 210 items), so the benchmark can be filtered by how independently each label is attested.
On the public set, mean discrimination---the score difference between the upper and lower halves of a 24-model analysis subset ranked by overall score---is positive for all four tasks, from $+0.16$ to $+0.24$.

\section{Evaluating Frontier Models}
\label{sec:eval}

\subsection{Protocol}
Prompts provide the scene, cast, target utterance, and question (\texttt{P\_context}). MCQs use JSON answers, exact-match scoring, and majority voting over three option permutations. Qwen3.8-Max scores T3 reactions against the gold descriptions on a 0/1/2 scale, with temperature 0 and thinking disabled (Appendix~\ref{app:judge}). Each T3 item averages its listener scores and divides by two; the overall score averages all 791 items. Controlled reasoning comparisons fix the model version and vary its reasoning setting (Appendix~\ref{app:revision-eval}).

\subsection{Leaderboard}
GPT-6 Astra leads Table~\ref{tab:leaderboard} with 73.48 overall, followed by Gemini 3.7 Flash (72.28). Kimi K3 is the highest-scoring open-weight model at 66.76. DeepSeek V4 Pro scores highest on speaker intent (T1), while GPT-6 Astra leads on listener judgments (T2/T3). GPT-6 Astra and Gemini 3.7 Flash tie on politeness classification (T4), and Gemini also leads the MCQ tasks with 78.97. Predicting each listener's reaction in free text remains difficult, with the best T3 score reaching only 58.37.

Using Qwen3.8-Max and Claude Sonnet-4.6 as alternative T3 judges yields highly similar model rankings across 34 configurations, with Spearman correlations of $\JudgeTThreeRho$ for T3 and $\JudgeOverallRho$ for overall scores. Appendix~\ref{app:addeval} reports detailed cross-judge results.

\textbf{Training gains.} CPR raises Qwen3-8B's MCQ accuracy from 36.75 to 67.01 and its overall score from 34.20 to 56.69. Its T3 score is close to Base (27.39 versus 26.94). The training analysis in \S\ref{sec:training} examines how annotation traces and RL contribute to the MCQ gains.

\textbf{Reasoning budgets.} Increasing reasoning effort does not consistently improve mock-politeness understanding. DeepSeek V4 Pro improves from 58.36 overall with thinking off to 68.44 at max effort. Qwen3.8-Max performs best with thinking off (63.43), while Qwen3.8-27B peaks at low effort (54.64). The benefits of additional reasoning are model-dependent rather than uniformly positive (Appendix~\ref{app:revision-eval}).

\textbf{Character-name sensitivity.} On 508 quality-checked items, two consistent renamings change CPR's mean accuracy from 63.39 to 64.07, versus 36.22 to 32.48 for Base and 67.91 to 58.17 for Qwen3.8-Max. The intervention changes names and aliases while preserving dialogue and role relations (Appendix~\ref{app:revision-names}). CPR retains its accuracy under character renaming, whereas Base and Qwen3.8-Max are more sensitive to character-name cues.

\subsection{Text Ablations and Oracle Commentary}
\label{sec:ablation}
On the hidden set we vary the prompt for three models (base 8B, our trained model, Gemma-4-31B). Knowledge-enhanced prompting, which injects the textbook definition of mock politeness and helped in prior work \citep{maskofcivility2026}, hurts all three models here, by $-11.5$, $-15.8$, and $-4.6$ points respectively (Table~\ref{tab:ablation}), with verbatim prompts in Appendix~\ref{app:prompts}. Removing the cast block barely matters. Adding full scene context raises accuracy by 3.7 points for Gemma and 6.5 points for CPR, while lowering Base accuracy by 3.3 points. Appending a short excerpt of the expert's commentary (an oracle hint) lifts Gemma-4-31B to 85.2, the base model to 60.6, and our model to 75.5. The commentary supplies the expert's interpretation of the target utterance, including its implied intent and listeners' understanding. Subtitle-cropped frames yield no significant gain over mismatched frames; audio helps MiniCPM-o-2.6 ($+3.1$) and leaves Gemini-2.5-Pro unchanged (Appendix~\ref{app:vision}).

\begin{table}[t]
\centering
\small
\setlength{\tabcolsep}{3.1pt}
\begin{tabular}{lccc}
\toprule
\rowcolor{headrow} Model & Consist. & Per-align.\ mean & None right \\
\midrule
Kimi-K2.6 & \textbf{42.9} & \textbf{65.5} & \textbf{13.5} \\
Gemma-4-31B & 39.7 & 63.7 & 14.1 \\
Sonnet-4.6 & 36.5 & 60.4 & 16.7 \\
GPT-5 & 27.6 & 53.2 & 21.2 \\
Haiku-4.5 & 19.9 & 42.2 & 34.6 \\
Qwen3-8B (base) & 12.2 & 35.4 & 41.0 \\
\midrule
\rowcolor{refrow} \method{} (ours) & 37.2 & 64.1 & 10.9 \\
\bottomrule
\end{tabular}

\caption{Schema-enabled diagnostics over 156 alignments with at least two MCQ items (T1/T2/T4; percent). ``Consist.'' counts alignments with all sibling MCQs correct, ``None right'' with all wrong. The shaded row is our trained 8B model (\S\ref{sec:training}).}
\label{tab:diag}
\end{table}
\subsection{Schema-Enabled Diagnostics}
\label{sec:diag}
Because multiple tasks share an alignment, we can measure whether a model understands a scene coherently. Cross-task consistency (all sibling MCQs correct) separates models far more sharply than accuracy (Table~\ref{tab:diag}). For Kimi-K2.6 versus Base 8B, MCQ accuracy differs by 1.8$\times$ (66.8/36.8), compared with 3.5$\times$ for consistency (42.9/12.2); GPT-5 and Haiku-4.5 score 27.6 and 19.9. Coherent scene-level understanding is rarer than per-item accuracy suggests. The trained model raises consistency from 12.2 to 37.2 (above Sonnet-4.6) and reduces the none-right rate from 41.0 to 10.9, the lowest value in Table~\ref{tab:diag}. Mismatch-confusion analysis appears in Appendix~\ref{app:addeval}.

\section{Training Study}
\label{sec:training}

Understanding mock politeness requires relating polite wording to speaker intent and listener uptake in context. Our annotations record these relations as intent, emotion--language mismatch, and listener reactions, providing supervision for training. As RL gains saturate, \method{} uses a chance-corrected recovery estimate to select and weight existing gold traces for further supervision.

Unless noted otherwise, training uses Qwen3-8B with LoRA \citep{hu2022lora}; Appendix~\ref{app:addbackbone} replicates the supervised stage, one GRPO round, and one full loop iteration on further backbones. Unless noted, we evaluate on the MCQ items of the held-out hidden test, which shares no episode with training, using paired bootstrap tests. The public-set results use the MCQ protocol of Table~\ref{tab:leaderboard}, reported as the micro average over its MCQ tasks (T1, T2, T4); the overall column of Table~\ref{tab:leaderboard} additionally includes judge-scored T3.

\subsection{\agvr{}: Annotation-Grounded Verifiable Rewards}
\label{sec:agvr}

For SFT, \agvr{} converts each aligned annotation into a five-step trace: evidence, mismatch, per-listener uptake, intent, and a JSON answer. Each step is instantiated programmatically from gold fields. For RL, the same fields define string-verifiable rewards (answer, mismatch, and uptake matches plus a $-0.5$ format gate). Training rewards are computed directly from the gold annotation fields. Supervision lifts the 8B model from 36.8 to 52.3 on this metric and one GRPO round on the same annotations to 55.7, above GPT-5 (53.3); and on the 253 items the teacher Kimi-K2.6 could itself solve, gold traces beat its rejection-sampled chain-of-thought 55.6 to 49.5. Auxiliary reward heads reduce variability across seeds (Appendix~\ref{app:addtrain}). SFT accuracy increases with gold-trace count (Figure~\ref{fig:scaling}, Appendix~\ref{app:addtrain}).

\begin{figure}[t]
\centering
\includegraphics[width=\columnwidth]{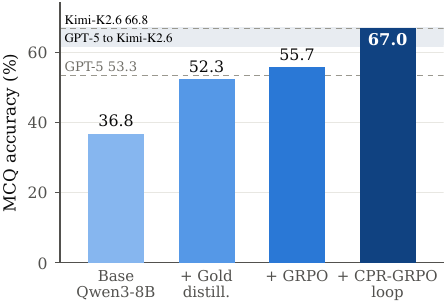}

\caption{Leaderboard-protocol MCQ accuracy (public set). Training raises 8B accuracy from 36.8 to 67.0; the shaded band spans GPT-5 to Kimi-K2.6.}
\label{fig:fairproto}
\vspace{-0.2cm}
\end{figure}

\subsection{Effects of Output Format and Supervised Content}
\label{sec:interface}

GRPO from the base model changes accuracy by $-0.5$ points ($p{=}0.9$). A two-by-two study (Table~\ref{tab:interface}, Appendix~\ref{app:addtrain}) separates the interface (the five-step-plus-JSON output format that the reward function parses and scores) from the content (gold reasoning). Interface-only SFT replaces reasoning spans with placeholders and reaches 56.9/55.6; subsequent GRPO adds $+3.3$/$+3.2$. Gold reasoning supplied as unparseable prose scores 1.9/3.7, with 95 percent parse failures and 51 percent zero-variance rollout groups. Interface-only SFT thus enables GRPO gains in both tested seeds. A shuffled-content control gains $+2.8$, compared with $+25.5$ for the full recipe. Answer-plus-intent supervision without the five-step structure scores 59.3 versus 62.0 (Table~\ref{tab:attr}). These controls identify contributions from correct answer supervision and a parseable reward interface.

\begin{table}[t]
\centering
\small
\setlength{\tabcolsep}{4pt}
\begin{tabular}{lcc}
\toprule
\rowcolor{headrow} Cold-start variant & Score & Paired $\Delta$ \\
\midrule
Gold scaffold (full recipe) & \textbf{62.0} & $+25.5$ \\
Answer + intent, no structure & 59.3 & $+22.7$ \\
Gold, matched 253 items & 55.6 & $+18.1$ \\
Teacher CoT, same 253 items & 49.5 & $+10.6$ \\
Shuffled content & 40.8 & $+2.8$ \\
\bottomrule
\end{tabular}
\caption{SFT-stage attribution on the hidden set (percent; paired $\Delta$ against the base of the same evaluation session). Content drives the gain; gold beats teacher CoT on identical items.}
\label{tab:attr}
\end{table}
\subsection{Chance-Corrected Analysis of GRPO}
\label{sec:chance}

\begin{figure}[t]
\centering
\includegraphics[width=\columnwidth]{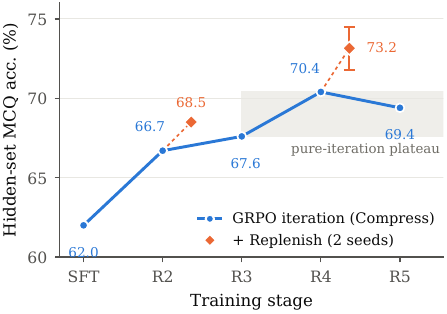}
\caption{Pure GRPO plateaus near 70 on the hidden set (shaded band); adding Replenish (orange, two seeds) reaches 74.5 at round 4.}
\label{fig:curve}
\vspace{-0.2cm}
\end{figure}

GRPO after SFT adds $+4$ to $+5$ points to greedy accuracy. pass@8 stays flat (93.1 to 93.5) while greedy and majority vote rise, replicating the sharpening view of RLVR \citep{yue2025does}. The pass@16-minus-greedy gap is 31.4 points; random guessing already gives pass@16 above 99 percent on 4-way items. Chance-corrected, the recoverable excess is only 4.2 points (RL) and 4.8 (SFT), a 7.5$\times$ inflation that declines monotonically down the pipeline (5.7, 4.8, 4.2) even as greedy accuracy climbs thirty points. The recoverable band decreases during training. Grouping greedy-wrong items by chance-corrected recovery shows that fewer items remain recoverable as GRPO gains diminish; the round-3 increment is $+0.9$.

\subsection{\method{}: Compress, Probe, Replenish}
\label{sec:cpr}

\begin{figure}[t]
\centering
\includegraphics[width=\columnwidth]{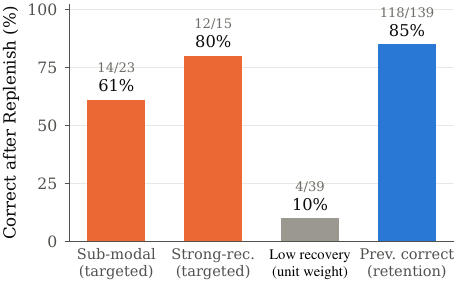}

\caption{After Replenish, correct-answer rates are highest among recoverable errors (orange) and lower among low-recovery errors (gray); most previously correct answers are retained (blue).}
\label{fig:buckets}
\end{figure}

\method{} uses the recoverable band to allocate further supervision from the expert-grounded gold traces through three steps (Algorithm~\ref{alg:cpr} and Figure~\ref{fig:loop}, Appendix~\ref{app:training}). \emph{Compress} runs one token-level GRPO round (asymmetric clipping 0.2/0.28, zero-variance filtering, k3 KL to the round-start policy; any RLVR optimizer fits this slot). \emph{Probe} re-samples the training set 16 times per item and buckets by chance-corrected recovery. \emph{Replenish} oversamples gold traces of sub-modal and strong-recovery items ($4\times$ and $2\times$) for one epoch of hard cross-entropy.

\noindent\textbf{Reward, objective, and probe.} Formally, let $\mathcal{D}=\{x_i\}$ be the training items, each with $k_i$ options and gold fields $(a_i^{*}, m_i^{*}, u_i^{*})$ for answer, mismatch label, and per-listener uptake, and let $\mathcal{G}=\{\tau_i\}$ be the programmatic gold traces of \S\ref{sec:agvr}. The verifiable reward of a rollout $y$ is
\begin{equation}
\label{eq:reward}
\begin{aligned}
r(y \mid x_i) ={}& \mathds{1}[\mathrm{ans}(y){=}a_i^{*}] + \lambda\,\mathds{1}[\mathrm{mis}(y){=}m_i^{*}] \\
&+ \lambda\,\mathrm{F}_{\mathrm{aud}}(y, u_i^{*}) - \tfrac{1}{2}\,\mathds{1}[\mathrm{malf}(y)],
\end{aligned}
\end{equation}
with $\lambda{=}0.3$, $\mathrm{F}_{\mathrm{aud}}$ the fraction of listeners whose uptake tag matches gold, and $\mathrm{malf}(\cdot)$ flagging malformed outputs. Within a group of $G$ rollouts the advantage is $\hat{A}_g = (r_g - \mu_i)/(\sigma_i + \epsilon)$, zero-variance groups are dropped, and the Compress step maximizes a token-level clipped objective with asymmetric clipping and a k3 KL anchor to the round-start policy (Eq.~\ref{eq:grpo}, Appendix~\ref{app:training}). The Probe step estimates, for every item, the empirical correct rate $\hat{p}_i$ over $G'{=}16$ fresh samples and the greedy correctness $g_i$, and converts them into a chance-corrected recovery excess and a replenishment weight
\begin{equation}
\label{eq:probe}
\begin{gathered}
e_i = \max\!\big(0,\; \hat{p}_i - \tfrac{1}{k_i}\big),\\
w_i =
\begin{cases}
4 & g_i{=}0,\, e_i{>}\delta,\, \hat{p}_i{\le}\tfrac{1}{2}\ \text{(sub-modal)}\\
2 & g_i{=}0,\, \hat{p}_i{>}\tfrac{1}{2}\ \text{(strong recovery)}\\
1 & \text{otherwise,}
\end{cases}
\end{gathered}
\end{equation}
with margin $\delta{=}0.06$. Items with $e_i \approx 0$ get no extra budget, separating the probe from classical hard-example mining. Appendix~\ref{app:addloop} reports sensitivity to the weights and margin. The Replenish step takes one epoch of weighted cross-entropy on gold traces, $\mathcal{L}(\theta) = -\sum_i w_i \log \pi_\theta(\tau_i \mid x_i)$, with no KL anchor to overwrite confidently wrong modes.

\noindent\textbf{Gains from replenishment.} Pure GRPO plateaus near 70 (rounds 2 to 5 give 66.7, 67.6, 70.4, 69.4), whereas Replenish reaches 68.5/68.5 at round 2 and 71.8/74.5 at round 4 (Figure~\ref{fig:curve}). Repeated evaluations give 74.5--77.3 before and 75.5--75.9 after another Compress step (Appendix~\ref{app:addtrain}). Figure~\ref{fig:buckets} reports changes within the probe-defined groups. Replenish flips 61 and 80 percent of the sub-modal and strong-recovery items against 10 percent of the unrecoverable bucket, keeps 85 percent of previously correct ones, and re-probing identifies 81 recoverable items at round 4, compared with 146 at round 2. The same targeting reproduces on two further backbones (Appendix~\ref{app:addbackbone}). Further Compress gains have paired intervals spanning zero. Replenish lowers scores from 75.9 to 70.4 and 69.4 to 68.1, with 1 and 8 sub-modal items remaining (Appendix~\ref{app:addloop}). Under the leaderboard protocol the best checkpoint scores 67.0 on the MCQ tasks (Figure~\ref{fig:fairproto}), above Sonnet-4.6 (61.9) and level with Kimi-K2.6 (66.8).

\subsection{Training Controls}
\label{sec:controls}

Table~\ref{tab:controls} (Appendix~\ref{app:addloop}) reports the training controls. Anchoring the same gold traces inside RL groups as off-policy members, in the spirit of LUFFY \citep{luffy2025}, yields 65.3, compared with 66.7 for GRPO. Pure GRPO scores 69.4 at round 5. Rejection sampling adds $+1.0$, DPO at $\beta{=}0.05$ reduces accuracy by 8.3 points, and five SFT epochs score 61.6 versus 62.0 after three epochs. CPR-GRPO achieves the highest accuracy among these training strategies.

\par
\begingroup
\subsection{Transfer to External Tasks}
\label{sec:revision-transfer}
We test both the fixed \bench{} checkpoints and the training recipe on external data.

\paragraph{Fixed-checkpoint transfer.}
Without target-data training, CPR exceeds Base, SFT, and GRPO in all six evaluations in Table~\ref{tab:revision-fixed}. CPR reaches 47.73 Macro-F1 on ToSarcasm and 86.00 accuracy on iSarcasmEval C, versus 42.79 and 82.00 for Base. Appendix~\ref{app:revision-fixed} details the datasets and evaluation protocols.

\begin{table}[!htbp]
\centering\small\color{black}
\setlength{\tabcolsep}{4.2pt}
\renewcommand{\arraystretch}{1.08}
\begin{tabular}{lrrr>{\columncolor{refrow}}r}
\toprule
\rowcolor{headrow} Dataset & Base & SFT & GRPO & \textbf{CPR} \\
\midrule
MultiPragEval (Zh) & 65.78 & 64.00 & 65.78 & \textbf{66.33} \\
MultiPragEval (En) & 65.56 & 65.89 & 65.78 & \textbf{67.11} \\
ToSarcasm & 42.79 & 46.68 & 45.93 & \textbf{47.73} \\
\quad without topic & 45.90 & 51.34 & 50.84 & \textbf{51.89} \\
iSarcasmEval A & 28.07 & 28.96 & 30.88 & \textbf{31.40} \\
iSarcasmEval C & 82.00 & 83.50 & 85.50 & \textbf{86.00} \\
\bottomrule
\end{tabular}

\caption{Transfer without target-data training. Scores are accuracy, except ToSarcasm (Macro-F1) and iSarcasmEval A (sarcastic-class F1). Details in Appendix~\ref{app:revision-fixed}.}
\label{tab:revision-fixed}
\vspace{-0.4cm}
\end{table}

\paragraph{Retraining on external datasets.}
We independently train Qwen3-8B on FGVIrony, CPED, and SemEval-2018 Task~3B with three seeds. Table~\ref{tab:revision-retraining} compares Base, SFT, GRPO, CPR, and wrong-only replenishment. CPR exceeds GRPO by 2.33, 3.03, 7.50, and 4.10 Macro-F1 points across the four settings, and also exceeds SFT and wrong-only replenishment. Appendix~\ref{app:revision-transfer} gives the data splits and training settings.

\begin{table}[!htbp]
\centering\small\color{black}
\setlength{\tabcolsep}{4.5pt}
\renewcommand{\arraystretch}{1.08}
\begin{tabular}{lrrrr}
\toprule
\rowcolor{headrow} Method & FGV & CPED-B & CPED-N & SemEval \\
\midrule
Base & 17.77 & 23.97 & 23.97 & 19.05 \\
SFT & 32.09 & 28.75 & 31.61 & 38.80 \\
GRPO & 38.19 & 29.07 & 32.48 & 43.16 \\
Wrong-only & 32.59 & 28.19 & 37.36 & 41.17 \\
\midrule
\rowcolor{refrow} \textbf{CPR} & \textbf{40.52} & \textbf{32.10} & \textbf{39.98} & \textbf{47.26} \\
\bottomrule
\end{tabular}

\caption{External retraining: mean Macro-F1 over three seeds for trained methods; Base uses one fixed checkpoint. B/N: balanced/natural CPED sampling. Details in Appendix~\ref{app:revision-transfer}.}
\label{tab:revision-retraining}
\vspace{-0.4cm}
\end{table}

\par
\endgroup

\section{Conclusion}
\label{sec:conclusion}

\bench{} evaluates mock politeness through speaker intent, emotion--language mismatch, and listener uptake. GPT-6 Astra leads the leaderboard with 73.48 overall, while listener-reaction prediction remains difficult across models. Gold annotation traces outperform teacher reasoning on matched items, and interface-only supervision enables subsequent GRPO gains. Chance correction reduces the estimated recoverable performance gap by 7.5$\times$; \method{} uses this estimate to weight gold supervision and raises an 8B model's public MCQ accuracy from 36.8 to 67.0. Fixed-checkpoint evaluation shows gains on Chinese and English sarcasm tasks, and character renaming preserves CPR's aggregate accuracy. Larger reasoning budgets improve DeepSeek's performance; Qwen models peak at off or low effort. Together, these findings show how expert pragmatic interpretations can support both the evaluation and training of models to understand mock politeness.

\section*{Limitations}
The benchmark draws its scenes from a single Chinese drama. Training and hidden splits use different episodes but share characters and setting. Overlap with model pretraining data is unknown. T3 is scored by an LLM judge. Its absolute scores vary across judges.

\section*{Ethics Statement}
\bench{} draws on publicly accessible, copyrighted creative works: a television screenplay, fan commentary, a recap podcast, and a novel, cited in \S\ref{sec:pipeline}. The release contains structured annotations, short evidence quotations, and episode and scene references; it excludes the original video and audio. The commentary author's account is cited, and the reception statistics in Appendix~\ref{app:schema} come from public profile pages. The dialogue and characters are fictional. Items contain indirect verbal aggression characteristic of mock politeness, with no slurs or graphic material. Potential misuse includes assistance with manipulative writing. Our experiments train and evaluate comprehension.

\bibliography{custom}

\clearpage
\appendix

\begin{center}
   {\Large \bf Supplementary Material \par}
\end{center}
\vspace{1.5em}

\etocdepthtag.toc{mtappendix}
\etocsettagdepth{mtmain}{none}
\etocsettagdepth{mtappendix}{subsection}
\renewcommand{\contentsname}{Appendix Contents}
\setcounter{tocdepth}{2}
\tableofcontents
\clearpage

\numberwithin{equation}{section}
\numberwithin{figure}{section}
\numberwithin{table}{section}
\renewcommand{\theHequation}{\thesection.\arabic{equation}}
\renewcommand{\theHfigure}{\thesection.\arabic{figure}}
\renewcommand{\theHtable}{\thesection.\arabic{table}}

\section{Dataset Construction and Statistics}
\label{app:schema}
\label{app:stats}

\begin{table}[t]
\centering
\small
\setlength{\tabcolsep}{2.7pt}
\begin{tabular}{lccccc}
\toprule
\rowcolor{headrow} Benchmark & Size & Tasks & ToM & Listener & Train. \\
\midrule
SwordsmanImp & 200 & 1 & -- & -- & -- \\
Mask of Civility & 100 & 1 & -- & -- & -- \\
PUB (En) & 28k & 14 & partial & -- & -- \\
EmotionQueen (En) & 10k & 4 & -- & -- & -- \\
\midrule
\rowcolor{refrow} \bench{} & \textbf{1{,}592} & \textbf{4} & \checkmark & \checkmark & \checkmark \\
\bottomrule
\end{tabular}
\caption{Pragmatics benchmarks. ``Listener'' marks per-listener uptake structure; ``Train.'' marks annotations designed to double as training signal.}
\label{tab:prior}
\vspace{-0.4cm}
\end{table}

Table~\ref{tab:stats} gives the per-task-by-split item counts and Table~\ref{tab:funnel} the construction funnel from raw alignments to final items. The paragraphs below document the annotation schema, the extraction and audit passes, and the assembly checklist behind those counts.

\begin{table}[h]
\centering
\small
\setlength{\tabcolsep}{5pt}
\begin{tabular}{lccccc}
\toprule
\rowcolor{headrow} Split & T1 & T2 & T3 & T4 & Total \\
\midrule
Public & 239 & 240 & 206 & 106 & 791 \\
Hidden test & 92 & 82 & 103 & 42 & 319 \\
Training & 140 & 158 & 120 & 64 & 482 \\
\midrule
All & 471 & 480 & 429 & 212 & 1{,}592 \\
\bottomrule
\end{tabular}

\caption{\bench{} items by task and split. Aligning 76 episodes (18{,}297 utterances, 1{,}214 scenes) with the 110-part expert commentary (47 hours) yields 1{,}506 alignments. The 68 podcast transcripts and 1.26M-character novel provide cross-source corroboration. Training and hidden episodes are disjoint.}
\label{tab:stats}
\end{table}

\paragraph{Annotation schema.} Each of the 1{,}506 alignments pairs one screenplay utterance with the expert's reading of it and carries four structured fields. \texttt{real\_intent} is a one-sentence statement of what the speaker is doing (e.g., \zh{借道贺暗指皇上极少召幸, 公开质疑龙种真伪}, ``using a congratulation to hint that the Emperor rarely summons her, publicly casting doubt on whether the child is genuine,'' for the running example of Figure~\ref{fig:overview}). \texttt{emotion\_language\_mismatch} takes one of \emph{aligned}, \emph{partial\_mismatch}, or \emph{inverted}, operationalizing the implicational mismatch account of mock politeness. \texttt{third\_party\_reactions} lists every listener present with an uptake tag in \{\emph{yes}, \emph{no}, \emph{partial}\} and a one-sentence reason. \texttt{rhetoric\_sub\_type} records the rhetorical device and supports the Culpeper three-way classification.

\paragraph{Extraction and audit.} Extraction runs in two passes. The first pass converts a commentary span into candidate fields with a verbatim quote anchor from the commentary, and a second audit pass verifies each field against its anchor and repairs or discards unsupported fields. On the full set the audit flagged 292 of the 1{,}506 alignments (19.4 percent), the leading error types being plot summary written as intent (136), character misattribution (62), hallucinated detail (61), and event misattribution (33), and repaired 269 of them against their anchors.

\paragraph{Item assembly.} Items are assembled under an anti-leakage checklist. Questions use neutral phrasing that does not presuppose hostility, scenes are truncated before the consequences of the utterance unfold, options are length-balanced within 15 percent, every MCQ is evaluated under three option permutations, and sibling items derived from one alignment never share option text.

\begin{table}[t]
\centering
\small
\setlength{\tabcolsep}{5pt}
\begin{tabular}{lr}
\toprule
\rowcolor{headrow} Stage & Count \\
\midrule
Aligned annotations & 1{,}506 \\
\quad flagged by audit pass (repaired) & 292 (269) \\
\midrule
Candidate items (two assembly passes) & 8{,}544 \\
\quad rejected: option/gold length balance & 4{,}471 \\
\quad rejected: options-only probe$^{\dagger}$ & 1{,}816 \\
\quad rejected: anti-leakage and consistency & 665 \\
\midrule
Final items (18.6\% of candidates) & \textbf{1{,}592} \\
\bottomrule
\end{tabular}
\caption{Construction funnel. $^{\dagger}$still solved from options alone after two rewrites. Anti-leakage rejections include circular evidence and future-tense narration.}
\label{tab:funnel}
\end{table}

\paragraph{Primary-source selection.} We selected the commentary for its line-by-line analysis of dialogue and quotable evidence for the annotated interpretations. The account is certified by Bilibili as a well-known film-and-television creator and has about 491{,}000 followers; the 47-video archive of parts 1--110 used in our corpus has 32.8 million plays (accessed 2026-07-31). Each extracted field carries a verbatim quote anchor and undergoes an audit against the commentary. The English recap podcast and source novel provide cross-source corroboration (\S\ref{sec:quality}).

\section{Evaluation Prompts}
\label{app:prompts}

\begin{table}[t]
\centering
\small
\setlength{\tabcolsep}{2.4pt}
\begin{tabular}{lccc}
\toprule
\rowcolor{headrow} Prompt condition & Base 8B & \method{} & Gemma-31B \\
\midrule
Dialogue only & 44.0 & 52.8 & 62.0 \\
+ Full scene (main) & 40.7 & 59.3 & 65.7 \\
\quad $-$ cast block & 41.2 & 58.3 & 66.2 \\
+ Knowledge (defn.) & 29.2 & 43.5 & 61.1 \\
+ Oracle commentary & \textbf{60.6} & \textbf{75.5} & \textbf{85.2} \\
\bottomrule
\end{tabular}

\caption{Text ablations on the hidden set (percent). The injected definition lowers all three scores; expert commentary yields the highest accuracy for each model.}
\label{tab:ablation}
\vspace{-0.2cm}
\end{table}

The main condition \texttt{P\_context} presents, in order, the cast block (name, rank, and a one-line persona for every character present), the scene header, the full scene dialogue up to the truncation point, the target utterance, the question, the options, and a JSON answer contract. \texttt{P\_lite} drops the cast block and scene. The knowledge-enhanced condition \texttt{P\_definition} prepends the following definition, given verbatim with an English gloss.

\begin{quote}\small
\zh{【Haugh 2014 mock politeness 定义 (理论参考)】mock politeness 指外形上礼貌 (敬语/谦辞/客套) 但内核为非礼貌 (攻击/讽刺/施压) 的语言现象。形式与功能错位, 用礼貌形式包装非礼貌意图。识别要点 (1) 表面话语寄存器为礼貌 (2) 真实意图为攻击或操控 (3) 听者通常听得懂, 但因规范无法当场揭破。}
\end{quote}

\begin{quote}\small
English gloss. [Haugh 2014 mock politeness definition, theoretical reference] Mock politeness is speech whose form is polite (honorifics, deference, courtesy formulas) but whose substance is impolite (attack, sarcasm, coercion). Form and function are misaligned, a polite wrapper around an impolite intent. Cues are (1) a polite surface register, (2) a real intent to attack or manipulate, (3) listeners who usually understand but cannot expose the attack on the spot.
\end{quote}

The oracle condition appends a block labeled \zh{【专家解说提示】} (``expert commentary hint'') containing the expert's commentary excerpt for the aligned utterance. The scaffold prompt used for training and hidden-set evaluation appends a five-step output contract requiring \texttt{<evidence>}, \texttt{<mismatch>}, \texttt{<listeners>}, and \texttt{<intent>} tags before the JSON answer. These prompt comparisons disable thinking. Reasoning-setting experiments use the configurations in Appendix~\ref{app:revision-eval}. RL-trained checkpoints repeated prompt text when evaluated with a thinking template absent from training.

\section{Training Details}
\label{app:training}

\begin{algorithm}[t]
\small
\caption{One \method{} iteration}
\label{alg:cpr}
\begin{algorithmic}[1]
\REQUIRE policy $\pi_\theta$ after gold-trace SFT; items $\mathcal{D}$ with gold $(a^{*}, m^{*}, u^{*})$; gold traces $\mathcal{G}$; group size $G$; probe size $G'$; margin $\delta$
\STATE \COMMENT{\textbf{Compress} (one GRPO round)}
\STATE $\pi_{0} \leftarrow \pi_\theta$; \ for each $x_i$: sample $G$ rollouts from $\pi_{0}$, score with Eq.~\ref{eq:reward}
\STATE drop zero-variance groups; $\hat{A}_g \leftarrow (r_g - \mu_i)/(\sigma_i + \epsilon)$
\STATE update $\theta$ by maximizing the clipped GRPO objective (Eq.~\ref{eq:grpo})
\STATE \COMMENT{\textbf{Probe} (chance-corrected band diagnosis)}
\STATE for each $x_i$: greedy correctness $g_i$; $\hat{p}_i \leftarrow$ correct rate over $G'$ samples at $T{=}0.8$
\STATE compute excess $e_i$ and weights $w_i$ by Eq.~\ref{eq:probe}
\STATE \COMMENT{\textbf{Replenish} (gold-targeted distillation)}
\STATE one epoch of weighted CE: $\mathcal{L}(\theta) = -\sum_i w_i \log \pi_\theta(\tau_i \mid x_i)$ \ (no KL anchor)
\STATE Return the updated policy.
\end{algorithmic}
\end{algorithm}

\begin{figure}[t]
\centering
\includegraphics[width=\columnwidth]{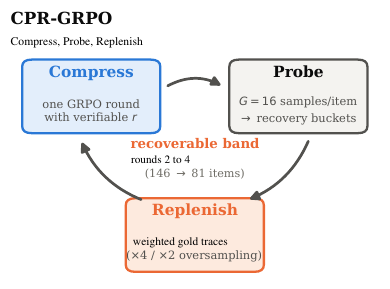}

\caption{The \method{} loop. Compress runs GRPO, Probe estimates chance-corrected recovery, and Replenish weights gold traces using those estimates.}
\label{fig:loop}
\end{figure}

All main runs use Qwen3-8B with LoRA rank 64, $\alpha{=}128$, dropout 0.05, on all seven projection matrices (174.6M trainable parameters, 2.1 percent), on a single H200 GPU. SFT over 338 traces, one GRPO round, one probe, and one Replenish epoch each complete within hours; a full loop iteration fits within a single GPU-day. SFT uses learning rate $10^{-5}$, three epochs over 338 traces. Each GRPO round samples $G{=}6$ rollouts per item at temperature 0.8 from the round-start policy, drops zero-variance groups, and trains one epoch with token-level importance ratios, asymmetric clipping $(\varepsilon_l, \varepsilon_h) = (0.2, 0.28)$, and a k3 KL penalty ($\beta{=}0.03$) to the round-start policy $\pi_{0}$, which acts as a per-round reference reset. The full Compress objective is
\begin{equation}
\label{eq:grpo}
\begin{aligned}
\mathcal{J}(\theta) ={}& \mathbb{E}\Big[ \tfrac{1}{|y|}\textstyle\sum_{t} \min\big( \rho_t \hat{A},\; \bar{\rho}_t\, \hat{A} \big) \Big] \\
&- \beta\, \mathrm{D}_{k3}\big(\pi_\theta \,\|\, \pi_{0}\big),
\end{aligned}
\end{equation}
where $\rho_t = \pi_\theta(y_t \mid \cdot)/\pi_{0}(y_t \mid \cdot)$ is the token importance ratio and $\bar{\rho}_t = \mathrm{clip}\big(\rho_t,\, 1{-}\varepsilon_{l},\, 1{+}\varepsilon_{h}\big)$ its clipped counterpart. Rewards follow Eq.~\ref{eq:reward} with $\lambda{=}0.3$. The probe uses $G'{=}16$ samples at temperature 0.8, margin $\delta{=}0.06$, and the Replenish step is one epoch of hard cross-entropy with oversampling weights 4 and 2 and no KL anchor. With sequence-level importance ratios, the KL term diverged to above $3{\times}10^{4}$ within one epoch, degrading the policy below its SFT starting point. Token-level ratios with the same data trained stably (final policy loss $3{\times}10^{-3}$, KL $\approx 0.009$).

\section{T3 Judge}
\label{app:judge}

T3 responses are scored per listener using the following rubric. The leaderboard and reasoning comparisons use Qwen3.8-Max, temperature 0, thinking disabled, and a 256-token output cap. The judge receives the scene, speaker, target utterance, listener, reference reaction, and predicted reaction. Model identities and previous scores are omitted.

\begin{quote}\small
\zh{【任务】对比 模型输出的 listener 反应描述 vs gold listener 反应描述, 评分 0/1/2。【评分细则】0 分: 模型反应与 gold 主旨明显不同/完全没答到点/方向相反。1 分: 部分正确, 抓到了 gold 的某一面但缺关键细节或有小错。2 分: 与 gold 主旨一致, 关键判断 (听懂/没懂/反击/顺从) 都对, 用词可以不同。}
\end{quote}

\begin{quote}\small
English gloss. [Task] Compare the model's description of each listener's reaction against the gold description and score 0/1/2. [Rubric] 0, the response clearly departs from the gold reading, misses the point entirely, or points the opposite way. 1, partially correct, capturing one aspect of the gold but missing key details or making a minor error. 2, consistent with the gold reading, with the key judgments (understood or not, strikes back or complies) all correct; wording may differ.
\end{quote}

Scores are averaged over listeners and normalized to $[0,1]$ per item. The judge appears only at evaluation time and never in training.

\section{Multimodal Probes: an Audit Protocol}
\label{app:vision}

We test whether frames from the scene help models read mock politeness. Multimodal social-interaction work reports that visual and auditory cues carry signal in multi-party scenes \citep{kang2025can,ouyang2025multi}, motivating an evaluation of visual context for mock-politeness understanding. Each probe item pairs a benchmark MCQ with six frames spanning $[t_0{-}2\mathrm{s}, t_0{+}6\mathrm{s}]$ around the utterance moment, located by bigram containment against Whisper segments. Three conditions are compared under identical text: \textbf{T} (text only), \textbf{TV} (text plus the item's own frames), and \textbf{VC} (text plus another item's frames under a fixed derangement), so TV$-$VC measures the difference between matched and mismatched visual context. Two frame sources are used: commentary-video frames (70 items) and episode frames (56 items), the latter also in a subtitle-cropped variant since episode frames carry hardcoded subtitles. Table~\ref{tab:vision} gives the full matrix.

\begin{table}[t]
\centering
\small
\setlength{\tabcolsep}{3.6pt}
\begin{tabular}{llcccc}
\toprule
\rowcolor{headrow} Model & Frames & T & TV & VC & TV$-$VC \\
\midrule
Qwen3-VL-8B & comm. & 46.7 & 51.0 & 51.0 & $+0.0$ \\
Qwen3-VL-8B & epis. & 40.5 & 44.6 & 43.5 & $+1.2$ \\
Qwen3-VL-8B & crop & 39.9 & 44.6 & 43.5 & $+1.1$ \\
Sonnet-4.6 & epis. & 58.3 & 60.7 & 65.5 & $-4.8^{*}$ \\
Sonnet-4.6 & crop & 58.3 & 62.5 & 62.5 & $+0.0$ \\
Gemini-2.5-Pro & epis. & 67.9 & 70.8 & 74.4 & $-3.6$ \\
Gemini-2.5-Pro & crop & 71.4 & 73.2 & 75.6 & $-2.4$ \\
\bottomrule
\end{tabular}
\caption{Vision probe (percent; paired). $^{*}$95\% CI $[-8.9, -0.6]$; all other TV$-$VC differences include zero.}
\label{tab:vision}
\end{table}

Both matched and mismatched frames improve accuracy over text alone, with TV$-$T gains of $+4$ to $+11$ points. On episode frames, Sonnet's TV$-$VC is $-4.8$ points (95\% CI $[-8.9,-0.6]$); after subtitle cropping, it is 0.0. All subtitle-cropped conditions in Table~\ref{tab:vision} have TV$-$VC confidence intervals that include zero.

\paragraph{Scaffolding and reaction-shot variants.} We vary visual response scaffolding and frame timing to test how models use visible expressions and listener reactions. We run a two-by-two on Qwen3-VL-30B-A3B over a 138-item probe set (414 paired observations per cell, all subtitle-cropped): frames at the utterance moment versus reaction shots at $[t_0{+}2\mathrm{s}, t_0{+}10\mathrm{s}]$, each answered directly versus with a describe-then-answer visual chain of thought that first describes every visible character's expression and body language. The mismatched-frame control is kept in every cell. TV$-$VC is $+0.7$ $[-1.4, +3.1]$, $+0.2$ $[-1.4, +1.9]$, $+0.5$ $[-1.9, +2.9]$, and $+1.0$ $[-1.2, +3.1]$ across the four cells, with all four confidence intervals including zero.

\paragraph{An audio probe.} We test whether prosodic cues improve performance using the same protocol on the 138-item set. Each item pairs its MCQ with the 8-second original-audio clip of the utterance moment (16 kHz mono, cut by the same Whisper timestamps), and three conditions are compared, \textbf{T} (text only), \textbf{TA} (text plus the item's own audio), and \textbf{AC} (text plus another item's audio under the same fixed derangement).

We compare matched and mismatched audio against the text-only baseline to distinguish their effects. On both MiniCPM-o-2.6 and Gemini-2.5-Pro \citep{comanici2025gemini}, matched audio beats the mismatched control by an identical TA$-$AC of $+3.6$, significantly so in both cases, yet the decomposition against text-only is opposite. For MiniCPM, matched audio raises accuracy over text (TA$-$T $=+3.1$ $[+0.2,+6.3]$) while a random clip is neutral (AC$-$T $=-0.5$). For Gemini, matched audio leaves accuracy unchanged (TA$-$T $=+0.0$ $[-3.1,+3.1]$) and mismatched audio lowers accuracy (AC$-$T $=-3.6$ $[-6.8,-0.5]$).

Qwen3-Omni-30B-A3B \citep{xu2025qwen3}, has negative point estimates for both audio comparisons (TA$-$T $=-2.9$, TA$-$AC $=-2.4$, both intervals spanning zero), and its visual comparison gives (TV$-$VC $=-0.2$). Matched-minus-mismatched differences reflect changes in either condition; comparison with text alone distinguishes improvements from losses under mismatched inputs. The same confound explains the vision numbers, where both matched and mismatched frames raise Gemini's accuracy over text (TV$-$T $=+4.3$, VC$-$T $=+6.8$) so TV$-$VC is negative. Audio improves MiniCPM in this evaluation; corresponding gains are absent for Gemini and Qwen3-Omni.

\section{Additional Results}
\label{app:additional}

\subsection{Evaluation Robustness}
\label{app:addeval}

\paragraph{T3 judge robustness.}
We score the same T3 responses with Qwen3.8-Max and Sonnet-4.6 under the rubric in Appendix~\ref{app:judge}. Across all 34 configurations, Spearman correlation is $\rho=\JudgeTThreeRho$ for T3 and $\rho=\JudgeOverallRho$ for overall scores (Table~\ref{tab:judge-comparison}). MCQ scores are fixed.

\begin{table*}[t]
\centering\small

\setlength{\tabcolsep}{4pt}
\begin{tabular}{lrrrr@{\hskip 12pt}lrrrr}
\toprule
Model & Q-T3 & S-T3 & Q-All & S-All & Model & Q-T3 & S-T3 & Q-All & S-All \\
\midrule
GPT-6 Astra & 58.37 & 61.08 & 73.48 & 74.19 & Qwen3-32B & 34.43 & 36.65 & 44.11 & 44.69 \\
Gemini 3.7 Flash & 53.28 & 54.37 & 72.28 & 72.57 & Qwen3.5-9B & 31.07 & 31.03 & 41.59 & 41.58 \\
Kimi K3 & 53.44 & 55.83 & 66.76 & 67.38 & Qwen3-14B & 30.91 & 32.40 & 39.15 & 39.54 \\
DeepSeek V4 Pro & 49.39 & 52.99 & 62.55 & 63.49 & GPT-5-mini & 35.68 & 35.88 & 38.12 & 38.17 \\
Qwen3.8-Max & 49.84 & 49.47 & 62.54 & 62.44 & Qwen3-VL-8B & 31.96 & 36.29 & 37.40 & 38.53 \\
Kimi-K2.6 & 48.18 & 46.20 & 61.98 & 61.46 & Qwen3-8B & 26.94 & 30.99 & 34.20 & 35.25 \\
Gemma-4-31B & 42.03 & 44.50 & 59.62 & 60.26 & GLM-4.6V-Flash & 28.80 & 33.45 & 33.16 & 34.38 \\
Sonnet-4.6 & 50.61 & 55.34 & 58.94 & 60.18 & Qwen3.5-4B & 15.82 & 16.91 & 32.94 & 33.23 \\
Gemini-3.1-Flash-Lite & 45.75 & 47.82 & 58.57 & 59.10 & InternVL3-8B & 26.01 & 28.64 & 31.81 & 32.49 \\
DeepSeek-V4-Flash & 45.27 & 44.09 & 56.80 & 56.49 & GLM-4.1V-9B & 25.81 & 29.25 & 31.12 & 32.02 \\
\method{} & 27.39 & 30.99 & 56.69 & 57.63 & GPT-4o-mini & 28.28 & 29.98 & 30.75 & 31.19 \\
Qwen3-VL-235B & 42.56 & 44.62 & 52.42 & 52.96 & Qwen3-VL-4B & 22.01 & 25.69 & 30.26 & 31.22 \\
GPT-5 & 49.47 & 51.09 & 52.33 & 52.75 & GLM-4.5-Air & 29.49 & 29.25 & 30.06 & 29.99 \\
GPT-4o & 38.96 & 41.63 & 49.34 & 50.03 & MiniCPM-o-2.6 & 21.72 & 25.16 & 29.42 & 30.32 \\
GLM-4.6V & 37.38 & 41.87 & 47.41 & 48.58 & MiMo-VL-7B & 28.72 & 29.69 & 28.34 & 28.59 \\
Qwen3-VL-30B & 36.89 & 41.14 & 44.37 & 45.48 & Qwen2.5-VL-7B & 22.57 & 22.90 & 28.26 & 28.34 \\
Haiku-4.5 & 40.29 & 43.61 & 44.37 & 45.24 & Qwen3.5-2B & 12.62 & 12.78 & 26.04 & 26.09 \\
\bottomrule
\end{tabular}
\caption{Cross-judge scores on identical responses (percent). Q: Qwen3.8-Max; S: Sonnet-4.6. MCQ scores are fixed. Both judges use the same rubric with thinking disabled.}
\label{tab:judge-comparison}
\end{table*}

\paragraph{Mismatch confusion.} On the hidden set, both trained models over-predict \emph{inverted} and struggle most on \emph{partial\_mismatch} (gold distribution 92/71/53 for inverted, partial, aligned; majority baseline 42.6). The SFT model classifies mismatch at 50.0 percent and the RL model at 50.9, with 42 of 71 partial items misread as inverted by the SFT model, indicating frequent confusion between partial and inverted mismatch.

\subsection{Training Analyses}
\label{app:addtrain}

\begin{table}[t]
\centering
\small
\setlength{\tabcolsep}{5pt}
\begin{tabular}{lccc}
\toprule
\rowcolor{headrow} Cold start & SFT & +GRPO & Observation \\
\midrule
Full recipe & 62.0/61.1 & 66.7/66.7 & baseline \\
Interface only & 56.9/55.6 & 60.2/58.8 & GRPO gain \\
Content, no iface & 1.9/3.7 & 5.6/25.0 & Parse failures \\
None (base) & --- & 40.3 & No gain \\
\bottomrule
\end{tabular}

\caption{Interface-alignment two-by-two on the hidden set, two seeds. Interface-only SFT enables GRPO gains in both seeds; unparseable gold prose yields low scores.}
\label{tab:interface}
\end{table}

\paragraph{Performance across prompts and tasks.} CPR scores 59.3 under the standard hidden prompt, 74.5 under the training-matched scaffold prompt, and 67.0 on public MCQs under the leaderboard protocol. The public-set score uses the leaderboard prompt in Table~\ref{tab:leaderboard}. CPR scores 27.39 on public T3, close to Base at 26.94. Training on gold \texttt{why} phrases monotonically hurts (34.3, 32.0, 28.5). The \texttt{real\_intent} field states the intended interpretation; \texttt{why} records the evidence supporting an uptake label. Training on the latter lowers reaction-prediction scores in these experiments.

\paragraph{Auxiliary rewards.} Across three independent runs, answer-only GRPO scores $65.1\pm2.8$ and the three-head mix scores $66.2\pm0.8$ (mean $\pm$ SD). Auxiliary rewards reduce variability across runs, with a smaller change in mean accuracy.

\begin{figure}[t]
\centering
\includegraphics[width=\columnwidth]{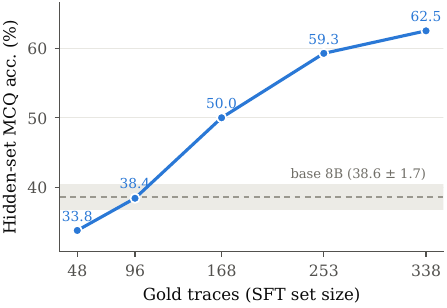}
\caption{AGVR SFT data efficiency on the hidden set. Accuracy crosses the base band (mean and SD of 62 evaluations) between 96 and 168 traces and saturates near 62.5.}
\label{fig:scaling}
\end{figure}

\paragraph{Data efficiency.} Figure~\ref{fig:scaling} varies the number of gold traces in AGVR SFT (nested subsets, three epochs, identical otherwise). The hidden-set score crosses the base between 96 and 168 traces and reaches 62.5 at the full 338, with clearly diminishing returns past 253 (a further 85 traces add 3.2 points). The supervised stage uses 338 human-anchored traces and requires no model-generated rationales.

\begin{figure}[t]
\centering
\includegraphics[width=\columnwidth]{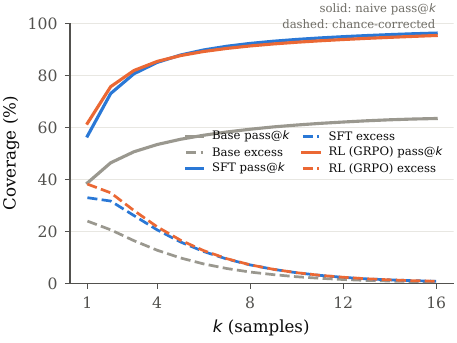}

\caption{Naive pass@$k$ (solid) vs.\ chance-corrected coverage (dashed). Chance-corrected coverage approaches zero at $k{=}16$.}
\label{fig:passk}
\end{figure}

\paragraph{Chance-Corrected pass@$k$.} Figure~\ref{fig:passk} plots naive pass@$k$ against its chance-corrected counterpart for the base, SFT, and RL checkpoints. At $k{=}16$, naive pass@$k$ approaches 95 percent while chance-corrected coverage approaches zero as random-guess coverage increases. The chance-corrected value at $k{=}1$ rises with training, from 23.9 at the base to 33.0 after SFT and 38.2 after RL. This is the full-curve form of the 7.5$\times$ inflation reported in \S\ref{sec:chance}.

\paragraph{Item-level flow.} Tracking each of the 216 hidden items across base, SFT, GRPO round 2, Replenish, and the loop peak shows the stages behave differently. SFT gains 67 items but overwrites 21 of the base model's 88 (76 percent retention); GRPO retains 92 percent of previously correct answers; Replenish retains 84 percent and corrects the most errors. Twenty-two items (10 percent) are never solved at any stage, indicating errors that persist through all evaluated stages, and 54 (25 percent) are solved at every stage.

\paragraph{Evaluation reproducibility.} Repeated greedy decoding of the same 216 hidden items produces checkpoint-dependent score variation. Base averages 38.6 with SD 1.7 over 62 evaluations (range 34.3--41.7); parse failures are at most 5.6 percent and are uncorrelated with score. SFT scores range from 58.8 to 62.0. Repeated scores are 66.2/66.2/66.7 for GRPO round 2, 75.5/75.9/75.9 for the additional Compress checkpoint, and 74.5/76.4/77.3 for the round-4 checkpoint. Paired bootstrap intervals compare predictions on the same items within an evaluation.

\subsection{\method{} Loop Ablations}
\label{app:addloop}

\paragraph{Additional uses of recovery estimates.} The band supports three uses, a hard filter (in-band groups at $G{=}16$ reach 68.1/71.3), a soft advantage weight (excess-weighted GRPO beats its uniform control on identical rollouts, unlike raw-variance weighting \citep{vcrl2025}), and the Replenish gate.

\begin{table}[t]
\centering
\small
\setlength{\tabcolsep}{3.5pt}
\begin{tabular}{lcc}
\toprule
\rowcolor{headrow} Control & Score & Factor tested \\
\midrule
GA-GRPO (gold in groups) & 65.3 & re-exposure \\
Round-5 pure iteration & 69.4 & more compute \\
RFT (rejection sampling) & 63.0 & self-distill \\
DPO ($\beta{=}0.05$) & 53.7 & preference opt. \\
More SFT (5 epochs) & 61.6 & more gradient \\
Ans-only vs.\ 3-head & 65.1/66.2 & reward richness \\
Teacher CoT (253 items) & 49.5/55.6 & teacher swap \\
\bottomrule
\end{tabular}
\caption{Controls on the hidden set. Vanilla GRPO from the same start scores 66.7; \method{} reaches 71.8/74.5 in two seeds.}
\label{tab:controls}
\vspace{-0.4cm}
\end{table}

\paragraph{Replenish constant sensitivity.} Rebuilding the Replenish set from the same round-2 diagnosis under alternative constants and retraining from the same checkpoint gives 67.6 for flattened weights 2/2, 65.7 for aggressive weights 8/4, and 63.0 for a tightened margin $\delta{=}0.12$, against 68.5 for the default 4/2/0.06. Equal weights remove the sub-modal emphasis, and the larger 8/4 weights lower accuracy. Tightening the margin reduces the selected sub-modal set from 29 items at $\delta{=}0.06$ to 20.

\paragraph{Excess-weighting dose response.} From the round-2 checkpoint, one sampling of 16 rollouts per item gives 68.5, 71.3, 73.1, 72.2, 69.9 for $\alpha=0,2,4,6,8$, an inverted-U peaking near $\alpha{=}4$. A second independent sampling gives 70.4, 73.1, 69.9 for $\alpha=0,2,4$. At $\alpha{=}2$, weighting improves over the corresponding uniform control by 2.8 and 2.7 points in the two samples. The two samples reach their highest scores at $\alpha{=}4$ and $\alpha{=}2$, respectively. Halving the probe to $G'{=}8$ shifts the diagnosed band only slightly (25 and 30 items against 29 and 33) and reaches 66.2 against 68.5 for the $G'{=}16$ default, a 2.3-point decrease with half as many probe samples.

\paragraph{Additional iteration.} From the 74.5 checkpoint, Compress reaches 75.9 (T1/T2/T4 of 75.0/76.8/76.2), the probe finds one sub-modal and 17 strong-recovery items with a residual excess of 10 item-equivalents on the training set, and an additional Replenish step yields 70.4. From the 68.5 checkpoint, Compress reaches 69.4, the probe finds 8 and 27 with 15 item-equivalents, and an additional Replenish step yields 68.1. Paired confidence intervals for both Compress increments include zero (see \emph{Evaluation reproducibility}, Appendix~\ref{app:addtrain}). Excess-weighted GRPO scores 69.0 and 75.5 for the additional Compress step, compared with 69.4 and 75.9 for unweighted GRPO. Earlier-round weighting results appear in \emph{Excess-weighting dose response} above.

\subsection{Cross-Backbone Replication}
\label{app:addbackbone}

We apply the same recipe to GLM-4-9B \citep{glm2024chatglm} and Qwen3-14B. On GLM-4-9B, AGVR SFT lifts the hidden-set score from 42.1 to 54.2 (paired $+12.0$, 95\% CI $[+4.2, +19.9]$, $p=0.004$). In the paired GRPO comparison, SFT scores 55.6 and one Compress round raises accuracy to 60.2 ($+4.6$). On Qwen3-14B the identical pipeline goes 37.0 to 70.4 after SFT (paired $+33.3$) and 71.3 after one GRPO round, so the 14B SFT endpoint already exceeds the 8B model's GRPO endpoint. GRPO adds $+4.6$, $+4.7$, and $+0.9$ points on GLM-4-9B, Qwen3-8B, and Qwen3-14B, respectively; the smallest gain occurs at the highest SFT endpoint. SFT and GRPO improve scores on all three tested backbones.

\paragraph{Full-loop transfer.} The complete Compress--Probe--Replenish iteration also transfers to both backbones. On GLM-4-9B, from the Compress checkpoint (58.3 in the paired evaluation), the probe diagnoses 37 sub-modal and 34 strong-recovery items ($\Sigma e = 21.4$), and one Replenish epoch lifts the same 216 items to 60.6 (paired $+2.3$, 95\% CI $[-5.1, +9.7]$), reducing the sub-modal set on re-probing (11 items, $\Sigma e = 13.6$). Flips concentrate where diagnosed, with 78 and 88 percent of sub-modal and strong-recovery training items flipping correct against 32 percent of the untargeted greedy-wrong bucket, at 91 percent retention. A second Compress round changes accuracy by $+4.2$ points (95\% CI $[-2.3, +10.6]$).

On Qwen3-14B the probe finds a smaller band (17 sub-modal and 19 strong-recovery items, $\Sigma e = 12.0$), yet one Replenish epoch raises hidden-set accuracy from 69.9 to 75.5 (paired $+5.6$, 95\% CI $[-0.5, +11.6]$), reducing the sub-modal set from 17 to 6 items while re-probing identifies further strong-recovery items. The targeting contrast is clear for strong-recovery items (84 percent flip correct) but weak for the already-small sub-modal bucket. A second Compress round changes the hidden score by $-2.3$ points (95\% CI $[-8.3, +3.7]$) and reduces the two recovery groups to 8 items each ($\Sigma e = 5.5$).

\paragraph{Recovery estimates across backbones.} Single-iteration Replenish gains are $+1.8$, $+2.3$, and $+5.6$ points on the three backbones. Both GLM-4-9B and Qwen3-14B gain at $\Sigma e \approx 12$, while the additional Compress step changes their scores by $+4.2$ and $-2.3$ points, respectively. These comparisons relate recovery estimates to the observed gains from further training.

\par
\begingroup
\section{Model Comparisons and Transfer}
\label{app:revision}

\subsection{Model Performance and Reasoning Settings}
\label{app:revision-eval}

\paragraph{Scoring.}
The leaderboard and reasoning comparisons use the 791 public items: 585 MCQs (T1/T2/T4) and 206 T3 items. MCQs use three option permutations and majority voting with the fixed tie rule. Qwen3.8-Max applies the rubric in Appendix~\ref{app:judge}. Overall score is $(585\,\mathrm{MCQ}+206\,\mathrm{T3})/791$. Invalid final answers receive zero.

\paragraph{Models and reasoning settings.}
Table~\ref{tab:leaderboard} spans 11 model families \citep{yang2025qwen3,qwen2025qwen3,bai2025qwen25vl,wang2024qwen2vl,kimiteam2025k2,deepseekai2024v3,deepmind2025gemma4,glm2024chatglm,hong2025glm,zhu2025internvl3,yao2024minicpmv,mimovl2025,openai2025gpt5,openai2024gpt4o,anthropic2025claudesonnet46,anthropic2025claudehaiku45,google2025gemini31pro}.
The reasoning comparison fixes each model version and the public items, varies the reasoning setting, and allows up to 32,768 output tokens. Temperature is zero where supported. Qwen3.8-27B runs locally in BF16. Each setting contributes one generation pass.

\begin{table}[t]

\centering\small
\setlength{\tabcolsep}{3.3pt}
\begin{tabular}{lrrrr}
\toprule
Model & Off & Low & High/Med. & Max/Xhigh \\
\midrule
DeepSeek V4 Pro & 58.36 & 61.13 & 65.53 & \textbf{68.44} \\
Qwen3.8-Max & \textbf{63.43} & 63.42 & 61.60 & 62.31 \\
Qwen3.8-27B & 51.57 & \textbf{54.64} & 54.09 & 52.13 \\
\bottomrule
\end{tabular}
\caption{Overall score across reasoning settings, with Qwen3.8-Max judging T3. DeepSeek uses off/low/high/max; Qwen uses off/low/medium/xhigh, with a 16K thinking budget for Qwen-Max medium.}
\label{tab:revision-reasoning}
\end{table}

DeepSeek gains 10.07 points from off to max (scene-bootstrap 95\% CI $[7.41,12.80]$). Qwen-Max's off and low settings score almost identically; all its enabled-versus-off confidence intervals include zero. Qwen-27B improves by 3.07 points from off to low ($[0.23,6.02]$), with smaller gains at higher budgets. Across the 12 configurations, overall-score rankings under Qwen and GLM-5.3 judging are identical (Spearman $\rho=1.00$).

\subsection{External Evaluation of Fixed Checkpoints}
\label{app:revision-fixed}

We evaluate the Qwen3-8B Base, SFT, GRPO, and CPR checkpoints without training on target-dataset labels. SFT is trained on 338 gold traces.

MultiPragEval \citep{park2024multiprag} contributes 300 Chinese and 300 English pragmatic-inference questions; accuracy averages three option permutations. ToSarcasm \citep{liang2022tosarcasm} uses its 973-item Chinese test set, evaluated both with the topic and after removing it, and reports binary Macro-F1. English iSarcasmEval \citep{abufarha2022isarcasm} uses 1,400 Task-A items (sarcastic-class F1) and 200 Task-C pairs (accuracy under the original order). Each pair contrasts a sarcastic text with its non-sarcastic rephrase.

Table~\ref{tab:revision-fixed} in \S\ref{sec:revision-transfer} reports the results. CPR improves over Base on all six evaluations, including $+4.94$ Macro-F1 on ToSarcasm and $+3.33$ sarcastic-class F1 on iSarcasmEval A. Removing the topic raises ToSarcasm scores for every checkpoint.

\subsection{Retraining on External Datasets}
\label{app:revision-transfer}

\paragraph{Tasks and splits.}
We independently train Qwen3-8B on seven-class Chinese FGVIrony \citep{wen2025fgvirony}, 19 dialogue acts from Chinese CPED \citep{chen2022cped}, and four-class English irony from SemEval-2018 Task~3B \citep{vanhee2018irony}. FGVIrony training examples sharing a news component with the test set are removed; development data are grouped by news component. CPED uses dialogue-disjoint warm-start and subsequent training pools, in balanced and natural-frequency regimes. Balanced sampling takes 24 warm-start and 48 pool examples per class; natural sampling keeps the same total sizes. Both regimes use the same full official test set. SemEval uses the official cleaned test inputs joined to gold labels by identifier.

\begin{table}[t]

\centering\small
\setlength{\tabcolsep}{4pt}
\begin{tabular}{lrrrr}
\toprule
Dataset & Warm & Pool & Dev & Test \\
\midrule
FGVIrony & 1,024 & 2,048 & 958 & 1,973 \\
CPED (each regime) & 456 & 912 & 1,687 & 27,438 \\
SemEval Task 3B & 192 & 384 & 767 & 784 \\
\bottomrule
\end{tabular}
\caption{External training and evaluation sizes. Warm and Pool are disjoint supervised-start and subsequent training sets.}
\label{tab:revision-splits}
\end{table}

\paragraph{Training.}
All runs use LoRA rank 64 and scaling 128, two warm-start SFT epochs, and learning rates $10^{-5}$ for SFT and $5\times10^{-6}$ for GRPO. Compress uses six sampled responses per item; Probe uses 16, with chance level $1/K$, excess threshold 0.06, and replenishment weights 4/2/1. Rewards check the gold class; malformed answers receive $-0.5$. FGVIrony runs three cycles and selects checkpoints on development Macro-F1. CPED and SemEval use one fixed cycle. Checkpoint selection is completed before test evaluation. Wrong-only replenishment samples greedy errors after Compress and matches CPR's supervised output-token budget. Evaluation uses greedy decoding with thinking disabled.

Table~\ref{tab:revision-retraining} in \S\ref{sec:revision-transfer} reports the three-seed means. CPR exceeds GRPO by 2.33, 3.03, 7.50, and 4.10 points across the four settings, and also exceeds wrong-only replenishment. Prioritizing recoverable errors is more effective here than sampling all greedy errors indiscriminately.

\subsection{Character Renaming}
\label{app:revision-names}

\paragraph{Construction and checks.}
We exclude 20 pilot scenes from the 585 public MCQs, leaving 525 candidates. Two consistent pseudonym maps replace names, aliases, and name-bearing titles throughout the inputs while retaining rank, relationships, dialogue, and option meanings. Programmatic checks verify reversible replacement. Kimi K3 and GLM-5.3 review semantic consistency without access to gold answers. Flagged replacements are corrected and rechecked; a coding assistant checks remaining flags against the source text. Seventeen unresolved or name-dependent items are excluded, yielding 508 items from 170 scenes. The final set is frozen before evaluation.

\paragraph{Evaluation.}
We evaluate the original and both renamed versions with thinking disabled, temperature zero, a 1,024-token output cap, and three option permutations. Accuracy uses the same strict JSON parser and majority aggregation for all versions. Table~\ref{tab:revision-names} reports Base, CPR, DeepSeek V4 Pro, Qwen3.8-Max, and Qwen3.8-27B; $\Delta$ is the mean of A and B minus the original score. Dialogue and plot information remain available, so the intervention measures sensitivity to character names.

\begin{table}[t]

\centering\small
\setlength{\tabcolsep}{3pt}
\begin{tabular}{lrrrrr}
\toprule
Model & Orig. & A & B & Mean & $\Delta$ \\
\midrule
Base & 36.22 & 32.09 & 32.87 & 32.48 & $-3.74$ \\
CPR & 63.39 & 64.17 & 63.98 & 64.07 & $+0.69$ \\
DeepSeek V4 Pro & 61.81 & 56.69 & 59.84 & 58.27 & $-3.54$ \\
Qwen3.8-Max & 67.91 & 58.07 & 58.27 & 58.17 & $-9.74$ \\
Qwen3.8-27B & 56.50 & 52.36 & 52.95 & 52.66 & $-3.84$ \\
\bottomrule
\end{tabular}
\caption{Accuracy under two character-renaming maps on the same 508 items. Changes are percentage points.}
\label{tab:revision-names}
\end{table}

CPR retains its aggregate accuracy under both mappings, whereas Base and the three larger comparison models lose 3.54--9.74 points on average. Qwen3.8-Max shows the largest name sensitivity in this comparison.
\par
\endgroup

\end{document}